\documentclass{article}
\usepackage{ijcai26}

\usepackage{times}
\usepackage{soul}
\usepackage{url}
\usepackage[hidelinks]{hyperref}
\usepackage[utf8]{inputenc}
\usepackage[small]{caption}
\usepackage{graphicx}

\usepackage{booktabs}
\usepackage{algorithm}
\usepackage{algorithmic}
\usepackage[switch]{lineno}

\usepackage{amsmath,amssymb,amsfonts}
\usepackage{algorithmic}
\usepackage{algorithm}
\usepackage{graphicx}
\usepackage{hyperref}
\usepackage{natbib}
\usepackage{booktabs}
\usepackage{xcolor}
\usepackage{subcaption}

\title{FlashbackCL: Mitigating Temporal Forgetting in Federated Learning}

\author{
    Mubarak A. Ojewale \and Adriana E. Chis \and Jorge M. Cortés-Mendoza \and Bernardo Pulido-Gaytan \and Horacio González-Vélez
    \affiliations
    Cloud Competency Centre, National College of Ireland, Dublin, Ireland
    \emails
    mubarak.ojewale@ncirl.ie, 
    adriana.chis@ncirl.ie,
    JorgeMario.CortesMendoza@ncirl.ie,
    LuisBernardo.PulidoGaytan@ncirl.ie,
    horacio@ncirl.ie
}

\date{}

\begin{document}

\maketitle

\begin{abstract}
Federated Learning (FL) of foundation and edge models increasingly targets deployments where client data distributions drift over time, yet existing forgetting-mitigation methods assume each client's distribution is stationary. Flashback, the strongest recent FL method against cross-client (spatial) forgetting, uses monotonically accumulating per-class label counts as a knowledge proxy; this proxy becomes miscalibrated under temporal distribution shift and anchors the global model to an outdated class balance. 
%We formalise \emph{temporal forgetting} in FL, introduce a per-phase metric that isolates it from protocol-level fluctuations, 
We formalise \emph{temporal forgetting} in FL with a per-phase metric isolated  from protocol-level fluctuations and propose Flashback Continual Learning (FlashbackCL), a drop-in extension of Flashback with $\mathit(i)$ temporally-decayed label counts; $\mathit(ii)$ a device-aware replay buffer with %class-balanced reservoir sampling
Class-Balanced Reservoir Sampling (CBRS); and $\mathit(iii)$ server-side active coreset curation on the public distillation set. %On CIFAR-10 with 50 clients and three controlled temporal shift modes, FlashbackCL reaches %55.7\% to 57.5\% final accuracy versus Flashback's 51.9\% to 52.3\% (6.9\% to 10.0\% relative improvement)
The results show that FlashbackCL achieves 6.9\% to 10.0\% relative improvement relative to Flashback, on CIFAR-10 with 50 clients and three controlled temporal shift modes, while simultaneously reducing temporal forgetting by up to 68\%. A 5-variant ablation identifies CBRS replay as the critical component. FlashbackCL also improves Flashback by 3.5 points on stationary CIFAR-100, suggesting that class-balanced replay regularises spatial heterogeneity as well as temporal shift. 
\end{abstract}

\section{Introduction}

Federated Learning (FL) has become the default paradigm for training shared models across decentralised clients without moving private data to a central server~\citep{mcmahan2017communication,kairouz2019advances}. Recent work on FL of foundation models and large edge models %~\citep{li2020federated,lin2020ensemble} 
has pushed the research agenda toward realistic deployment settings where clients are resource-constrained and their data distributions are strongly non-Independent and Identically Distributed (non-IID) ~\citep{li2020federated,lin2020ensemble}. A key insight from the last two years is that spatial data heterogeneity in FL can be re-cast as a \emph{forgetting} problem%~\citep{aljahdali2024flashback,lee2021preservation}
: local updates overwrite global knowledge that other clients have contributed, and coordinate-wise averaging destroys class-specific knowledge embedded in divergent client models ~\citep{aljahdali2024flashback,lee2021preservation}. \citet{aljahdali2024flashback} introduced Flashback, which uses per-class label counts as a knowledge proxy and combines client-side and server-side dynamic distillation weighted by those counts, yielding strong gains over proximal, contrastive, and not-true-distillation baselines.

Flashback and its predecessors rest on an implicit assumption: each client's local data distribution is stationary over time,
%. In real-world edge deployments the assumption fails. 
which fails in real-world edge deployments.
In Connected and Cooperative Autonomous Mobility (CCAM), for example, traffic patterns shift with time of day and season, weather alters sensor readings, and road layouts change. In FL of foundation models at the edge, device-side data tracks the evolving behaviour of users rather than a static sample. Each device's data distribution therefore evolves across FL rounds, introducing a temporal dimension to heterogeneity that existing methods do not address.

Flashback's knowledge proxy is particularly vulnerable to temporal shift because the label count vector accumulates monotonically across rounds: %once a class has been seen, its count never decreases. 
the count of seen classes never decreases. 
%When a class becomes rare or disappears from the federation's active data, the label count still assigns it high distillation weight,
The label count still assigns high distillation weights to classes even if they become rare or disappear from the federation’s active data, which anchors the global model to an outdated class balance and dampens adaptation.

\paragraph{Contributions.} 
%We (i) formalise temporal forgetting as a failure mode distinct from the spatial forgetting already studied, and propose a per-phase metric that isolates it; 
We $\mathit(i)$ formalise and propose an isolated per-phase metric for temporal forgetting as a failure mode distinct from the spatial forgetting already studied;
%(ii) introduce \emph{FlashbackCL}, a drop-in extension of Flashback with three components: 
$\mathit(ii)$ introduce \emph{FlashbackCL} as an extension of Flashback with three additional components:
a temporally-decayed label count to release pressure on vanished classes, a device-aware replay buffer admitted via %class-balanced reservoir sampling
Class-Balanced Reservoir Sampling
(CBRS) so that the buffer is not flushed at phase boundaries ~\citep{chrysakis2020online}, and server-side active coreset curation that re-weights the public distillation set using per-class trend estimates; and $\mathit(iii)$ report controlled experiments on CIFAR-10 dataset with 50 clients, and CIFAR-100 dataset with 100 clients, across three shift modes (abrupt, gradual, and cyclic). On CIFAR-10, FlashbackCL reaches from 55.7\% to 57.5\% final accuracy, versus Flashback's 51.9\% to 52.3\%, and reduces temporal forgetting by up to 68\%. In CIFAR-100, FlashbackCL improves over Flashback by 9.7\% to 12.7\% in all three modes. A 5-variant ablation isolates CBRS replay as the primary mechanism.
%EWC (mode collapse to a single predicted class under abrupt and cyclic shift).

\section{Background and Related Work}

\subsection{Federated Learning (FL)}

We consider the standard cross-device FL setup %~\citep{mcmahan2017communication,kairouz2019advances,bonawitz2019towards} 
with $N$ clients~\citep{mcmahan2017communication,kairouz2019advances,bonawitz2019towards}. A client $i$ holds a private dataset $\mathcal{D}_i = \{(x_j, y_j)\}_{j=1}^{n_i}$ and the goal is to train a single global model that minimises
{\small
\begin{equation}
\min_{w \in \mathbb{R}^d} \;\sum_{i=1}^{N} \frac{|\mathcal{D}_i|}{|\cup_{j} \mathcal{D}_j|} L_i(w),
\label{eq:fl_objective}
\end{equation}}%
where $L_i(w) = \tfrac{1}{|\mathcal{D}_i|}\sum_j \ell(w; (x_j, y_j))$ is the local loss for client $i$ and $\ell$ is typically cross-entropy. Federated Averaging (FedAvg)
%~\citep{mcmahan2017communication}
solves \eqref{eq:fl_objective} through partial participation: at each communication round $t$, the server selects $K$ clients $\mathcal{S}_t \subseteq [N]$, broadcasts the current global model $w_{t-1}$ to them, lets each one perform $E$ epochs of local Stochastic Gradient Descent (SGD), and then aggregates the returned models by sample-weighted averaging $w_t = \sum_{k \in \mathcal{S}_t} \tfrac{n_k}{\sum_j n_j}\, w_{k,t}$ ~\citep{mcmahan2017communication}.

\paragraph{Heterogeneity-aware FL.}
FedAvg degrades under non-IID client data because each client's local SGD pulls $w_{k,t}$ toward a different optimum, an effect that compounds with $E$ and with the spread of label distributions across clients. Three complementary families of methods address this issue. Optimisation-side approaches such as FedProx~\citep{li2020federated} and Stochastic Controlled Averaging for FL (SCAFFOLD)~\citep{karimireddy2020scaffold} keep local updates close to a global anchor, through a proximal term $\tfrac{\mu}{2}\|w - w_{t-1}\|^2$ in the local objective or through control variates that estimate and cancel client-specific drift. Representation-side approaches such as MOdel-cONtrastive learning (MOON)~\citep{li2021model} use contrastive losses to align local and global feature spaces, while matching-based aggregation~\citep{wang2020federated} reduces the destructive interference inherent in coordinate-wise averaging. Distillation-side approaches such as FL Distillation Fusion (FedDF)~\citep{lin2020ensemble} replace weighted averaging with ensemble knowledge distillation~\citep{hinton2015distilling} against a public dataset, treating each participating client as a teacher. \citet{wang2020tackling} formalised the objective inconsistency that local-epoch heterogeneity introduces, motivating per-client normalisation. All of these approaches assume that the only source of distributional change between rounds is which clients are sampled; none address \emph{temporal} drift inside any single client.

\subsection{Forgetting in Federated Learning}

\citet{lee2021preservation} introduced FedNTD, which masks the ground-truth class in the KL divergence so that local training preserves global knowledge on classes that the client has not just seen, and showed that this mitigation alone yields competitive accuracy with FedDF. \citet{xu2022acceleration} proposed FedReg, generating Fast Gradient Signed Method (FGSM) perturbed synthetic data from the global model to regularise local updates against forgetting ~\citep{goodfellow2014explaining}. \citet{dupuy2023quantifying} provided large-scale measurements of catastrophic forgetting in continual federated learning and showed that its magnitude scales with both spatial and temporal heterogeneity. \citet{aljahdali2024flashback} reframed FL convergence as forgetting at both the local-update and aggregation phases, introducing the round-forgetting metric $F_t = -\tfrac{1}{C}\sum_c \min(0, A_t^c - A_{t-1}^c)$, where $C$ is the number of classes and $A_t^c$ is the global model's accuracy on class $c$ at round $t$. This metric captures only negative changes in per-class accuracy and so isolates loss of knowledge from gain of knowledge in other classes.

\paragraph{Flashback's dynamic distillation.}
Flashback uses per-class label counts as a proxy for per-class model knowledge. Given a student label count $\boldsymbol{\nu} \in \mathbb{R}^C$ and teacher label counts $\{\boldsymbol{\mu}_i\}_{i=1}^M$, the per-class distillation weights are $\alpha_i^c = \mu_i^c / (\nu^c + \sum_k \mu_k^c)$ and the dynamic-distillation loss is
{\small
\begin{align}
\mathcal{L}_{\text{dKD}} = \alpha_s^y \mathcal{L}_{\text{CE}} + \sum_i \mathcal{L}_{\text{dKL}}(F_{w_i}, F_{w_s}; \alpha_i),
\label{eq:dkd}
\end{align}}%
with $\mathcal{L}_{\text{dKL}}(p,q;\alpha_i) = \sum_c \alpha_i^c \cdot p^c \log(p^c/q^c)$. Flashback applies \eqref{eq:dkd} in two places: at each client during local training (with the global model as the single teacher) and at the server during a public-dataset distillation step (with the participating clients' local models plus the previous global model as teachers). The global label count $\boldsymbol{\pi}$ is built incrementally by adding $\gamma \boldsymbol{\mu}_k$ for each participant $k$, gated by $\gamma r_k \leq 1$ where $r_k$ is the client's lifetime participation count.

System-side work on resource-efficient client selection is orthogonal to but interacts with our temporal-shift question, because client-selection skew can amplify drift in the global label count ~\citep{abdelmoniem2023refl,fourati2023filfl}.

\subsection{Continual Learning (CL)}

Continual Learning (CL) trains a model on a sequence of tasks without revisiting prior data~\citep{parisi2019continual,delange2021continual}. Three families of methods address catastrophic forgetting. Regularisation-based methods such as %Elastic Weight Consolidation
EWC~\citep{kirkpatrick2017overcoming} penalise changes to parameters important for previous tasks, where importance is quantified by the diagonal Fisher information matrix; Learning without Forgetting~\citep{li2017learning} distils from the old model to the new one. Replay-based methods~\citep{chaudhry2018riemannian} maintain a buffer of past examples and interleave them with current training data; the effectiveness depends critically on which examples to store. %Class-balanced reservoir sampling (CBRS)
CBRS~\citep{chrysakis2020online} is a distribution-agnostic admission policy designed for online %continual learning
CL under class imbalance; it evicts from the majority class when the buffer is full and applies within-class reservoir sampling otherwise. Architecture-based methods~\citep{yoon2020scalable} allocate separate parameters per task and avoid interference at the cost of poor scaling.

\subsection{Federated Continual Learning (FCL)}

Task-incremental Federated CL (FCL) methods such as FCL with Weighted Inter-client Transfer (FedWeIT)~\citep{yoon2021federated} decompose model parameters into a shared federated component and sparse task-specific components, requiring known task identity at inference time. Continual FL with Distillation (CFeD)~\citep{ma2022continual} introduces a surrogate dataset on which both clients and the server distil from previous models when new tasks arrive, mitigating inter-task forgetting at both endpoints. Generative replay (Federated Class-Incremental Learning - FedCIL~\citep{qi2023better}) sidesteps explicit replay buffers by training a generator on previous-task data and synthesising replay samples on demand, which is attractive when raw replay would violate retention policies. Global-Local Forgetting Compensation (GLFC)~\citep{dong2022federated} combines class-aware gradient compensation with semantic relation distillation, while federated orthogonal training~\citep{bakman2023federated} aggregates per-client activation subspaces and projects gradients to lie outside spans where prior tasks have been learned. \citet{wuerkaixi2024accurate} explicitly models heterogeneity in how different clients forget, using a per-client forgetting score to weight aggregation. All of these methods, with the partial exception of CFeD, assume \emph{task-incremental} setups in which task identity and task boundaries are known. Our setting is \emph{domain-incremental}: the task (e.g.\ classify into 10/100 categories) is fixed throughout training, there are no explicit boundaries, and the input distribution drifts continuously. This regime more closely matches %CCAM and FM-era
CCAM and 5G/6G-era (FM-era) edge deployments, where the data-generating process evolves but the predictive task does not change.

%\subsection{Position of This Work}

\textbf{This work}. Heterogeneity-aware FL methods (FedProx, SCAFFOLD, MOON, and FedDF) target spatial heterogeneity and assume client distributions are stationary in time. FL-forgetting methods (FedNTD, FedReg, and Flashback) reframe spatial heterogeneity as forgetting at the local-update and aggregation phases but inherit the same stationarity assumption. Task-incremental FCL methods (FedWeIT, CFeD, FedCIL, GLFC, and FOT) target temporal change but require explicit task boundaries and are evaluated against task-incremental benchmarks. The unfilled gap is \emph{domain-incremental} FL, in which the predictive task is fixed and the per-client distribution drifts continuously. This is the regime FlashbackCL targets, and it is also the regime in which FL of foundation models on edge devices will operate, because user behaviour and sensor environments rarely respect explicit task boundaries. EWC~\citep{kirkpatrick2017overcoming} regularises updates by a diagonal Fisher matrix. LwF~\citep{li2017learning} distills from the previous-task model. Experience replay~\citep{chaudhry2018riemannian} stores past examples; CBRS~\citep{chrysakis2020online} is a distribution-agnostic admission policy for imbalanced online streams. We use CBRS inside federated client buffers, a combination that to our knowledge has not been explored.

\section{Temporal Forgetting in FL}
\label{sec:tf}

\paragraph{Setup.}
We extend the standard FL setup with time-varying client distributions. Each client $i \in [N]$ holds a dataset $\mathcal{D}_i^{(t)}$ at round $t$, drawn from a distribution $P_i^{(t)}$ that may differ across rounds. Past data may have been discarded; we do not assume clients can revisit $\mathcal{D}_i^{(t')}$ for $t' < t$. The server samples $\mathcal{S}_t \subseteq [N]$ with $|\mathcal{S}_t| = K$ at each round and aggregates local updates into a global model $w_t$.

\paragraph{Two orthogonal heterogeneities.}
We can decompose the deviation of the empirical FL objective at round $t$ from a hypothetical IID-stationary objective into two components. Let $T$ denote the total number of training rounds, $\bar{P}^{(t)} = \tfrac{1}{N}\sum_i P_i^{(t)}$ the federation-wide marginal at round $t$, and $\bar{P} = \tfrac{1}{T}\sum_t \bar{P}^{(t)}$ the time-averaged marginal. For any test loss functional $\ell$,
{\small
\begin{align}
\underbrace{\mathbb{E}_{P_i^{(t)}}[\ell] - \mathbb{E}_{\bar{P}}[\ell]}_{\text{total deviation}} = \;
&\underbrace{\mathbb{E}_{P_i^{(t)}}[\ell] - \mathbb{E}_{\bar{P}^{(t)}}[\ell]}_{\text{spatial heterogeneity at }t} \nonumber\\
&{}+ \underbrace{\mathbb{E}_{\bar{P}^{(t)}}[\ell] - \mathbb{E}_{\bar{P}}[\ell]}_{\text{temporal drift of population mean}}.
\label{eq:decomp}
\end{align}}%

The first term is the within-round, between-client variation that classical heterogeneity-aware FL methods target~\citep{li2020federated,karimireddy2020scaffold,lin2020ensemble}. The second term is the across-round, within-population shift that is the focus of this paper. The two are orthogonal in the sense that a method can address one without addressing the other; FedProx is the canonical example of a method that targets the spatial term but is structurally insensitive to the temporal term.
% , which our experiments will confirm.

\paragraph{Measuring temporal forgetting.}
Flashback's round-forgetting metric $F_t$ collapses both terms of \eqref{eq:decomp} into a single scalar and so cannot attribute observed accuracy drops to either source. To isolate the temporal component, we divide the $T$ training rounds into $P$ phases, where phase $j \in \{1, \ldots, P\}$ corresponds to a distinct distributional regime, and define
\begin{equation}
\text{TF}(j,k) = A^{(j)}(t_j^{\text{end}}) - A^{(j)}(t_{j+k}^{\text{end}}),
\label{eq:tf}
\end{equation}
where $A^{(j)}(t)$ is the global model's accuracy on a held-out test set drawn from phase-$j$'s distribution at round $t$ and $t_j^{\text{end}}$ is the last round of phase $j$. Positive values indicate forgetting; negative values indicate backward transfer. We report the mean TF ($\overline{\text{TF}}$) over valid $(j,k)$ pairs. By construction, $\overline{\text{TF}}$ is invariant to spatial heterogeneity at any given round and responds only to changes in the global model's competence on a fixed phase distribution as training progresses through subsequent phases.

\paragraph{Why Flashback fails under shift.}
Flashback's global label count $\boldsymbol{\pi}$ accumulates each participating client's local count $\boldsymbol{\mu}_k$ scaled by $\gamma$, subject to the trust gate $\gamma r_k \leq 1$. In closed form, the label count for class $c$ at round $t$ is
\begin{equation}
\pi_t^c = \sum_{t' \leq t} \sum_{k \in \mathcal{S}_{t'}} \gamma \cdot \mu_k^c(t') \cdot \mathbf{1}[\gamma r_k(t') \leq 1],
\label{eq:flashback_acc}
\end{equation}
where $r_k(t')$ is client $k$'s participation count up to round $t'$ and $\mathbf{1}[\cdot]$ is the indicator function. This sum is monotonically non-decreasing in $t$. The dynamic $\alpha$ derived from $\boldsymbol{\pi}$ therefore reflects a time-integrated history of class exposure rather than the current federation marginal. Concretely, suppose class $c$ was active during phases $1, \ldots, j-1$ and disappeared from all clients' active data at the start of phase $j$. The label count $\pi^c$ at any round $t > t_j^{\text{end}}$ remains lower-bounded by the total contribution accumulated across the first $j-1$ phases, which is large. The dynamic $\alpha^c$ continues to place distillation weight on $c$, with two consequences: during local training the client is forced to preserve outdated knowledge at the expense of the current distribution; during server distillation the global model is pushed to maintain confidence on a class that no recent participant has seen. Stationary settings benefit from this anchoring because it stabilises the distillation against noise from partial client participation; non-stationary settings are actively harmed by it.

\paragraph{Design implication.}
The decomposition in \eqref{eq:decomp} suggests that a temporally-aware extension of Flashback should attack both the temporal-drift term in the global label count (so the distillation proxy reflects recent rather than historical exposure) and the temporal-drift term in the per-client effective data (so clients can replay vanished classes when they have momentarily disappeared from their own loaders). FlashbackCL's three components map onto this prescription: temporal decay addresses the first, the device-aware replay buffer addresses the second, and active coreset curation provides a population-level safety net through the public dataset. Each component is independently ablatable and reduces to vanilla Flashback in the limit of its hyperparameter (Section~\ref{sec:ablation}).

\section{FlashbackCL}

Flashback Continual Learning (FlashbackCL) extends Flashback with three components that each address one aspect of the anchoring problem. When all three are disabled FlashbackCL reduces exactly to Flashback.

\subsection{Temporally-Decayed Label Count}

At the start of every round, we apply a decay factor $\lambda \in (0, 1]$ to the global label count before adding the round's contributions:
\begin{equation}
\boldsymbol{\pi}_t = \lambda \cdot \boldsymbol{\pi}_{t-1} + \sum_{k \in \mathcal{S}_t} \gamma_k \boldsymbol{\mu}_k,
\label{eq:decayed}
\end{equation}
where $\gamma_k$ follows Flashback's trust-gated rule. The decay lets $\pi^c$ decrease for classes that are no longer present in recent clients' data. A high $\lambda$ is robust to noise from partial participation but slow to respond to shift; a low $\lambda$ is responsive but sensitive to client-selection randomness. Section~\ref{sec:sensitivity} finds $\lambda = 0.95$ optimal on CIFAR-100 with 200 rounds, with robust performance across $\lambda \in [0.95, 0.99]$.

\subsection{Device-Aware Replay Buffer with CBRS}
\label{sec:replay}

Each client $k$ maintains a replay buffer $\mathcal{B}_k$ of capacity
\begin{equation}
|\mathcal{B}_k| = \min(M_k, \lfloor \beta / f_k \rfloor),
\label{eq:buffer_size}
\end{equation}
where $M_k$ is the device's memory budget (in number of samples), $f_k = r_k / t$ is client $k$'s participation frequency, and $\beta$ is a system-wide budget parameter. Clients that participate rarely receive a larger buffer because they rely more heavily on their own replay between participations.

\paragraph{Why a global-count admission policy fails.}
A natural design is to admit by inverse global count, $p(x,y) \propto 1/(\pi^y + \epsilon)$. We implemented this and found it fails catastrophically at phase boundaries: classes that have just appeared have $\pi^y \approx 0$ and therefore infinite priority, flushing the buffer of all prior-phase samples within a few rounds. This is the opposite of the desired behaviour.

\paragraph{CBRS admission.}
We instead use class-balanced reservoir sampling~\citep{chrysakis2020online}. Let $n_y$ be the count of class-$y$ items currently in the buffer and $n_y^{\text{seen}}$ the total class-$y$ samples ever seen. Given a new sample $(x,y)$:
\begin{enumerate}
\setlength\itemsep{1pt}
\item If the buffer has free capacity, admit unconditionally.
\item Otherwise let $y^* = \arg\max_c n_c$ be the majority class.
\begin{itemize}
\setlength\itemsep{1pt}
\item If $y \neq y^*$: evict a random item of class $y^*$ and admit $(x,y)$.
\item If $y = y^*$: apply within-class reservoir sampling, admitting as a random replacement with probability $n_y / n_y^{\text{seen}}$.
\end{itemize}
\end{enumerate}

CBRS is distribution-agnostic. New classes can only take slots from already-overrepresented classes, never from minority classes at risk of being forgotten. During local training, replay batches are mixed with current-data batches at a fixed ratio $\rho \in [0,1]$, so the effective loss per step is $(1 - \rho) \mathcal{L}_{\text{dKD}}(\text{current}) + \rho\, \mathcal{L}_{\text{dKD}}(\text{replay})$. We use $\rho = 0.4$.

CBRS makes a simple invariant hold: the buffer's per-class histogram is approximately uniform over the classes the client has ever seen, regardless of the order in which classes arrive. A buffer that respects this invariant cannot be flushed of a phase-$(j-1)$ class merely because phase-$j$ samples have started arriving, because phase-$(j-1)$ classes are minority classes by construction and CBRS only ever evicts from the majority. The invariant fails for any admission policy that scores incoming samples by a quantity correlated with novelty, because novelty is monotonically larger for the most recently appearing classes.

\subsection{Server-Side Active Coreset Curation}

The server maintains a sliding window of the last $W$ rounds of aggregated label-count inflows $\boldsymbol{m}_{t'} = \sum_{k \in \mathcal{S}_{t'}} \boldsymbol{\mu}_k$, and fits a per-class linear regression to estimate the slope $s^c$ of class $c$'s inflow over the window. Absolute slopes bias toward high-baseline classes (a class dropping $1000 \to 500$ produces a larger slope than $10 \to 5$). We therefore normalise by the per-class mean inflow $\bar{m}^c$ to obtain a unit-free relative slope:
\begin{equation}
\tilde{s}^c = \frac{s^c}{\max(\bar{m}^c, 10^{-6})}, \quad \bar{m}^c = \frac{1}{W}\sum_{t'=t-W+1}^{t} m_{t'}^c.
\end{equation}

The public-distillation sampler is then re-weighted as
\begin{equation}
\omega^c = (1-\eta) \cdot \mathrm{softmax}(-\tilde{s}^c/\tau) + \eta / C,
\label{eq:coreset_w}
\end{equation}
where $\tau > 0$ is a temperature controlling how sharply the curator concentrates on declining classes and $\eta \in [0,1]$ is a uniform-floor coefficient that guarantees every class receives at least $\eta / C$ sampling weight, preventing the curator from starving any class under sharp shifts.
\subsection{Analysis}

\paragraph{Reduction to Flashback.} FlashbackCL becomes Flashback with $\lambda = 1$, $|\mathcal{B}_k| = 0$, and $\eta = 1$. The extensions are additive, which makes the ablation in Section~\ref{sec:ablation} well-defined.

\paragraph{Overhead.} Temporal decay is an element-wise multiplication per round. The replay buffer adds one forward-backward pass per local step on a 32-sample batch against the 64-sample current batch, approximately 50\% overhead. Coreset curation adds a linear regression over $W \times C$ per round, negligible against server distillation. Communication cost is identical to Flashback.

\section{Experiments}

\subsection{Setup}
\label{sec:setup}

\paragraph{Datasets.} We evaluate FlashbackCL on two image-classification benchmarks under controlled temporal shift. In both cases, a 2.5\% random sample of the training set is held as the server-side public dataset. 
%\textbf{CIFAR-10} (primary): 10 classes partitioned across $N=50$ clients with Dirichlet $\beta = 0.1$, $K=10$ clients per round, and $T = 200$ rounds. The 10 classes are divided into $P=5$ groups of 2, creating 5 temporal phases of 40 rounds each. 
Primary: \emph{CIFAR-10} with 10 classes partitioned across $N = 50$ clients, Dirichlet $\beta = 0.1, K = 10$ clients per round, and $T = 200$ rounds, where the 10 classes are divided into groups of 2 (P = 5), creating temporal phases of 40 rounds each.
%\textbf{CIFAR-100} (secondary, harder): 100 classes across $N=100$ clients with Dirichlet $\beta = 0.1$, $K=10$, and $T=200$ rounds. The 100 classes are divided into $P=5$ groups of 20. 
Secondary (harder): \emph{CIFAR-100} with 100 classes across $N = 100$ clients, Dirichlet $\beta = 0.1, K = 10$, and $T = 200$ rounds, where the 100 classes are divided into groups of 20 (P = 5). 
The experimental evaluation considers three temporal shift modes: Instant switch at phase boundaries (\emph{abrupt}), 10-round linear transition window (\emph{gradual}), and phases wrap modulo $P$; phase length halved to fit two cycles (\emph{cyclic}). Finally, stationary CIFAR-100 backward-compatibility with $T=150$ and no shift is also reported.

\paragraph{Model and baselines.} Our implementation evaluates a 2-layer CNN with ${\sim}580$K parameters, following the architecture of \citet{mcmahan2017communication} and Flashback~\citep{aljahdali2024flashback}. The baseline algorithms on CIFAR-10 are FedAvg, Flashback, and FlashbackCL with three seeds each. For CIFAR-100, we additionally compare FedProx ($\mu=0.01$), FedNTD ($\beta=1$ and $\tau=3$).

\paragraph{FlashbackCL hyperparameters.} An SGD with lr $=0.01$, momentum $0.9$, and weight decay $10^{-4}$, and a CNN with batch 64 and $E=3$ local epochs. Additionally, distillation temperature 3, decay $\lambda=0.97$, replay mix $\rho = 0.4$, replay batch 32, per-client memory 10\,MB with $\beta=200$, sliding window $W=10$, coreset temperature $\tau=1.0$, uniform floor $\eta=0.5$, and seeds $\{42, 123, 456\}$.

\paragraph{Metrics.} Global and per-phase test accuracy (Acc) at the final round, and average temporal forgetting ($\overline{\text{TF}}$), see Equation (\ref{eq:tf}).

\paragraph{Hardware.} All experiments were run on a single Apple M4 chip using the PyTorch MPS backend. No multi-GPU parallelism was used. Total compute across all experiments reported in this paper was approximately 350 M4-hours.

\subsection{Results}
\label{sec:results}

\subsubsection{CIFAR-10: Main Results}

\begin{figure}[t]
\centering
\includegraphics[scale=0.6]{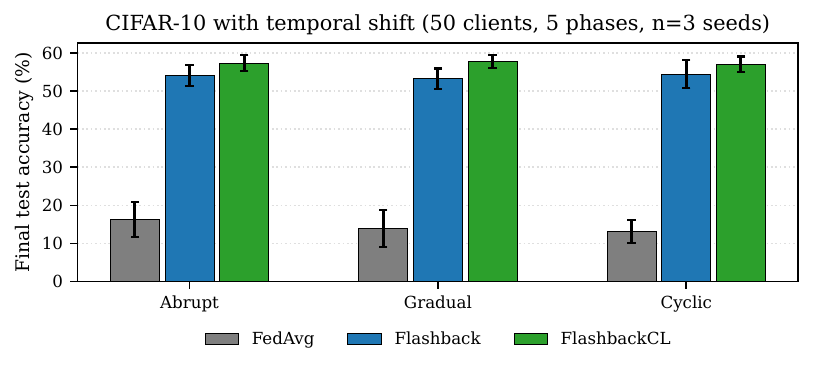}
\caption{Final test accuracy on CIFAR-10 with 50 clients across three temporal shift modes (mean over three seeds, error bars = standard deviation). FlashbackCL is the highest bar in every mode.}
\label{fig:cifar10_bars}
\end{figure}

\begin{table*}[t]
\centering
\small
\caption{CIFAR-10 with temporal shift: final test accuracy and temporal forgetting $\overline{\text{TF}}$, mean $\pm$ std over three seeds. Bold marks the best value per row. $\Delta_{\text{acc}}$ is the relative improvement of FlashbackCL over Flashback; $\Delta_{\text{TF}}$ is the absolute reduction in temporal forgetting.}
\label{tab:cifar10}
\begin{tabular}{l cc cc cc}
\toprule
& \multicolumn{2}{c}{FedAvg} & \multicolumn{2}{c}{Flashback} & \multicolumn{2}{c}{\textbf{FlashbackCL}} \\
\cmidrule(lr){2-3}\cmidrule(lr){4-5}\cmidrule(lr){6-7}
Shift & Acc & $\overline{\text{TF}}$ & Acc & $\overline{\text{TF}}$ & Acc ($\Delta_{\text{acc}}$) & $\overline{\text{TF}}$ ($\Delta_{\text{TF}}$) \\
\midrule
\emph{Abrupt}  & $0.157 \pm0.018$ & $0.179 \pm0.071$ & $0.519 \pm0.015$ & $0.126 \pm0.029$ & $\mathbf{0.566 \pm0.028}$\; ($+9.0\%$)  & $\mathbf{0.073 \pm0.041}$\; ($-42\%$) \\
\emph{Gradual} & $0.130 \pm0.052$ & $0.191 \pm0.076$ & $0.523 \pm0.019$ & $0.112 \pm0.015$ & $\mathbf{0.575 \pm0.028}$\; ($+10.0\%$) & $\mathbf{0.036 \pm0.011}$\; ($-68\%$) \\
\emph{Cyclic}  & $0.119 \pm0.029$ & $0.171 \pm0.156$ & $0.521 \pm0.038$ & $0.036 \pm0.079$ & $\mathbf{0.557 \pm0.019}$\; ($+6.9\%$)  & $0.065 \pm0.069$  \\
\bottomrule
\end{tabular}
\end{table*}

Table~\ref{tab:cifar10} and Figure~\ref{fig:cifar10_bars} report the main comparison on CIFAR-10. FlashbackCL reaches the highest final accuracy in every shift mode (55.7\% to 57.5\%), improving over Flashback by 6.9\% to 10.0\% relative. %Temporal forgetting is substantially reduced: on gradual \emph{shift}, FlashbackCL's $\overline{\text{TF}} = 0.036$ is 68\% lower than Flashback's $0.112$, and on abrupt shift 42\% lower.
FlashbackCL substantially reduces $\overline{\text{TF}}$ from $0.112$ to $0.036$, which represents a 68\% lower in \emph{gradual} shift, and 42\% lower for \emph{abrupt} shift than Flashback.
In particular, FlashbackCL also exhibits lower seed-to-seed variance, $\pm 0.019$ versus $\pm 0.038$, than Flashback with \emph{cyclic} shift, consistent with the ablation finding that the entire configuration stabilizes across runs (see Section~\ref{sec:ablation}). On cyclic shift, the $\overline{\text{TF}}$ comparison is not meaningful as Flashback's TF is already near zero ($0.036 \pm 0.079$) because phase cycling revisits old classes, and FlashbackCL's slightly higher value ($0.065 \pm 0.069$) reflects seed variance rather than a real degradation; the wide standard deviations overlap entirely. FlashbackCL also exhibits lower seed-to-seed variance than Flashback on cyclic shift ($\pm 0.019$ versus $\pm 0.038$), consistent with the ablation finding (Section~\ref{sec:ablation}) that the full configuration stabilises across runs.

% \color{red}
% Figure \ref{fig:cifar10_per_phase} shows...???
% \color{black}

% \begin{figure}[t]
% \centering
% \includegraphics[width=\linewidth]{figures/fig_cifar10_per_phase.pdf}
% \caption{CIFAR-10 \emph{abrupt}: per-phase accuracy trajectories (mean over 3 seeds). FlashbackCL retains higher accuracy on prior phases after new ones come online.}
% \label{fig:cifar10_per_phase}
% \end{figure}

\subsection{Cross-Dataset Replication: CIFAR-100}

\begin{figure}[t]
\centering
\includegraphics[width=\linewidth]{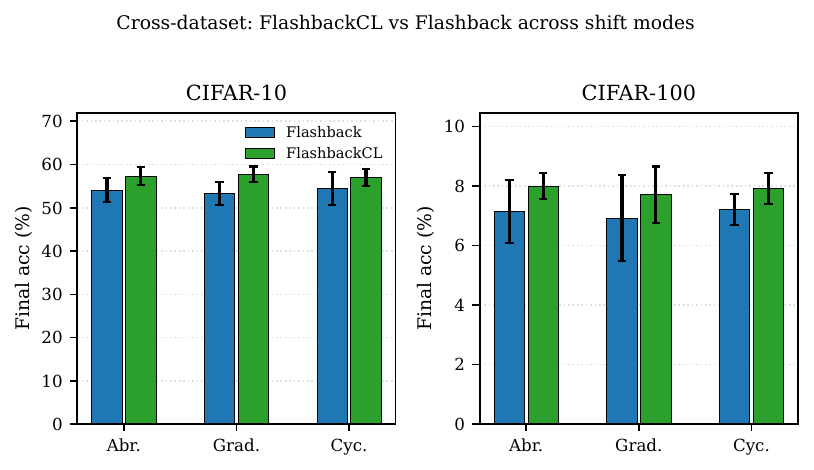}
\caption{Cross-dataset comparison: FlashbackCL (green) vs Flashback (blue) on CIFAR-10 (left) and CIFAR-100 (right) across three shift modes. The relative advantage replicates at both dataset scales despite the 10$\times$ difference in number of classes and clients.}
\label{fig:cross_dataset}
\end{figure}

\begin{table}[t]
\centering
\small
\caption{CIFAR-100 with temporal shift (100 clients, 200 rounds, $n=3$ seeds). FedProx tracks FedAvg closely, confirming that the proximal term addresses spatial but not temporal heterogeneity.}
\label{tab:cifar100}
\resizebox{\columnwidth}{!}{%
\begin{tabular}{lccccc}
\toprule
Shift & FedAvg & FedProx & FedNTD & Flashback & \textbf{FlashbackCL} \\
\midrule
Abrupt  & $.047 \pm .010$ & $.047 \pm .011$ & $.065 \pm .016$ & $.071 \pm .013$ & $\mathbf{.080 \pm .005}$\\
Gradual & $.057 \pm .013$ & $.056 \pm .013$ & $.072 \pm .012$ & $.069 \pm .018$ & $\mathbf{.077 \pm .012}$\\
Cyclic  & $.038 \pm .009$ & $.038 \pm .013$ & $.056 \pm .013$ & $.072 \pm .006$ & $\mathbf{.079 \pm .006}$\\
\bottomrule
\end{tabular}
}
\end{table}

Table~\ref{tab:cifar100} and Figure~\ref{fig:cross_dataset} show that the CIFAR-10 result replicates on CIFAR-100. FlashbackCL improves over Flashback by 9.7\% to 12.7\% across all three shift modes. FedProx tracks FedAvg within seed noise across all modes, confirming that the proximal term targets spatial heterogeneity but is structurally insensitive to temporal shift. The CIFAR-100 absolute numbers (5\% to 8\%) are lower than CIFAR-10 because of the 10$\times$ larger output space versus the same 580K-parameter CNN, and because the 5-phase protocol means only 20\% of test classes correspond to currently-active training classes. The relative improvements are the meaningful comparison.

% \subsubsection{EWC-Fed Mode Collapse (CIFAR-100)}
% \label{sec:ewc}

% With CIFAR-100, the full baseline set includes EWC-Fed, which exhibits a qualitative failure mode under \emph{abrupt} and \emph{cyclical} shifts. The trained global model converges to predicting a single class for every test example. The per-class Acc vector in the final round is perfect for class 0, zero elsewhere ($[1.0, 0.0,\ldots,0.0]$), giving a global accuracy of exactly 1\%, which is a random chance on a 100-way head. The pattern reproduces across all three seeds in both \emph{abrupt} and \emph{cyclic} modes; under \emph{gradual} shift EWC-Fed learns normally ($0.057 \pm 0.010$). Each client's Fisher anchor, estimated on its own local data, becomes stale against the post-aggregation global model, so the EWC penalty pulls local updates back toward an irrelevant parameter configuration. Under \emph{abrupt} and \emph{cyclic} shifts, the cumulative effect prevents any client from moving the global model off its initial state. At $\lambda_{\text{EWC}} = 1$, the penalty is weak enough that EWC-Fed escapes collapse ($0.032 \pm 0.003$) but still trails Flashback by four points.

\subsection{Public-Data Independence}
\label{sec:nopub}

A recurring concern with Flashback-family methods is that their advantage may stem from access to a public dataset rather than from the distillation mechanism itself. To test this, we disable all server-side distillation (\texttt{server\_epochs}$=0$) so that neither FedAvg, Flashback, nor FlashbackCL uses the public dataset in any way. The only difference between the three methods is their client-side training procedure.

\begin{table}[t]
\centering
\small
\caption{CIFAR-10 \textbf{without any public data} (server distillation disabled, $n=3$ seeds). FlashbackCL's advantage comes entirely from its client-side mechanisms (temporal decay + CBRS replay), not from public-data access.}
\label{tab:nopub}
\begin{tabular}{lccc}
\toprule
Shift & FedAvg & Flashback & \textbf{FlashbackCL} \\
\midrule
Abrupt  & $15.7 \pm 1.8$ & $15.9 \pm 3.2$ & $\mathbf{39.0 \pm 1.5}$ \\
Gradual & $13.0 \pm 5.2$ & $18.4 \pm 4.4$ & $\mathbf{37.5 \pm 3.6}$ \\
Cyclic  & $11.9 \pm 2.9$ & $17.3 \pm 3.6$ & $\mathbf{42.1 \pm 3.8}$ \\
\bottomrule
\end{tabular}
\end{table}

Table~\ref{tab:nopub} reports the results. Without public data, Flashback collapses to FedAvg-level accuracy (15.9\% vs 15.7\% on abrupt): its local dynamic distillation against the global model adds nothing when the global model's label count is stale and no server distillation corrects it. FlashbackCL, by contrast, reaches 37.5\% to 42.1\% across all three shift modes, more than doubling the baseline. The entire gain comes from temporal decay and CBRS replay operating on each client's own private data.

This result establishes that FlashbackCL's core contribution is \emph{independent of public-data access}. When a public dataset is additionally available, server-side distillation adds a further ${\sim}17$ points on top (Table~\ref{tab:cifar10}), but the client-side mechanisms provide a standalone $+19$ to $+25$ point advantage that no other tested method achieves.

\subsubsection{Ablation Studies}
\label{sec:ablation}

\begin{figure}[t]
\centering
\includegraphics[width=\linewidth]{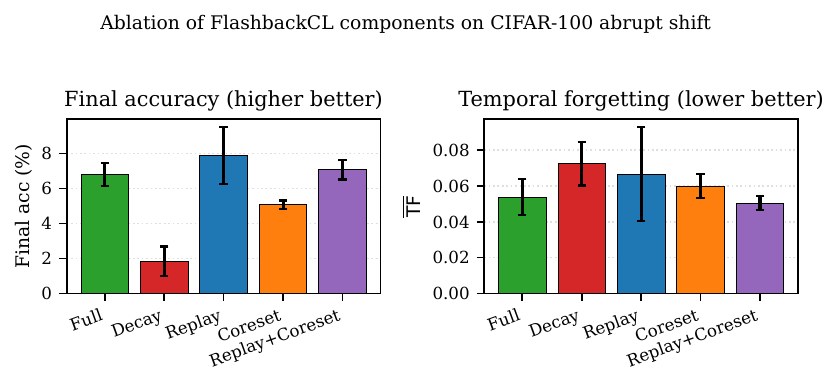}
\caption{5-variant ablation on CIFAR-100 abrupt (two seeds). Each variant has a distinct colour. Left: final accuracy (higher better). Right: temporal forgetting $\overline{\text{TF}}$ (lower better).}
\label{fig:ablation}
\end{figure}

Figure~\ref{fig:ablation} reports a 5-variant ablation of FlashbackCL on CIFAR-100 with \emph{abrupt} shift: \emph{decay}, \emph{replay}, \emph{coreset}, \emph{replay + coreset}, and \emph{full (decay + replay + coreset)}. The \emph{Full} variant is the original FlashbackCL version, while the other versions consider only some of its subcomponents. \emph{Replay} (CBRS) variants reach final accuracy within 1.2 points of the \emph{full} method, and variants without CBRS underperform vanilla Flashback; therefore, \emph{replay} is the critical component. The \emph{replay} version achieves the highest point with Acc = $0.079$, but has four times the variance of the \emph{full} configuration and worse temporal forgetting. The \emph{full} version trades a small amount of point Acc for substantially lower variance and lower $\overline{\text{TF}}$. 

Also, to isolate CBRS's contribution from \emph{replay} in general, we additionally compare against Replay-Fed, which uses standard random reservoir sampling (no class balancing) without distillation. On CIFAR-10 with \emph{abrupt} shift and $n=3$ seeds, Replay-Fed reaches $37.4 \pm 5.0\%$, well above FedAvg with $15.7\%$, but far below $51.9\%$ of Flashback and $56.6\%$ of FlashbackCL. This result confirms that (i) \emph{replay} variant without distillation cannot match distillation-based methods; and (ii) CBRS's class-balancing property is critical for the marginal gain over Flashback, since random eviction suffers from buffer flushing at phase boundaries as predicted by our design reasoning in Section~\ref{sec:replay}. Replay-Fed also exhibits twice the seed-to-seed variance of FlashbackCL ($\pm 5.0\%$ vs $\pm 2.8\%$), further validating the stability advantage of class-balanced admission.

\subsubsection{Decay-$\lambda$ Sensitivity and Backward Compatibility}
\label{sec:sensitivity}

\begin{figure}[t]
\centering
\includegraphics[width=0.9\linewidth]{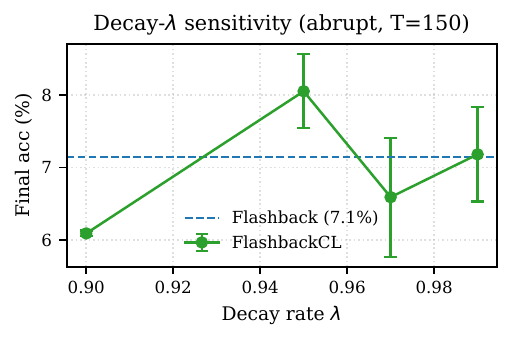}
\caption{Decay-$\lambda$ sensitivity for CIFAR-100 with \emph{abrupt} shift, $T=150$, and two seeds, and Flashback’s mean from the main results (dashed line).}
\label{fig:lambda_sens}
\end{figure}

Figure~\ref{fig:lambda_sens} sweeps the decay rate on CIFAR-100 with \emph{abrupt} shift. The optimum is $\lambda = 0.95$; performance is within seed noise across $\lambda \in [0.95, 0.99]$ and drops sharply at $\lambda = 0.90$ (over-decay). The main results use $\lambda = 0.97$, chosen before the sweep. We note that the optimal $\lambda$ value may be workload dependent.

\label{sec:backcompat}
On stationary CIFAR-100 with no shift and $T=150$, FlashbackCL reaches a 3.5-point improvement even without temporal shift than Flashback, from $0.265 \pm 0.007$ to $0.230 \pm 0.011$. Both trail FedAvg ($0.308 \pm 0.008$) because dynamic distillation adds noise when there is no forgetting to combat. This backward-compatibility result shows that CBRS replay is a useful regulariser for spatial heterogeneity in its own right.

\subsubsection{ResNet-18 Preliminary Result}
\label{sec:resnet}

To test the generalisability of CBRS as a primary finding beyond the 580K-parameter CNN, we performed a single-seed proof-of-scale experiment using ResNet-18 with ${\sim}11$M parameters on CIFAR-100 with \emph{abrupt} shift, $K=5$, and $T=100$ rounds. The ranking is maintained according to Table \ref{tb3} despite the reduced protocol. FlashbackCL's relative improvement over Flashback is +33\%. Evaluation of ResNet-18 with 3 seeds and 3 modes requires cluster computation, which is deferred to the extended version, but this single-seed result provides evidence that the temporal-forgetting advantage is not an artefact of the small-model regime.

\begin{table}[t]
\centering
\small
\caption{ResNet-18 with CIFAR-100, \emph{abrupt} shift, a single seed, $K{=}5$, and $T{=}100$.}
\label{tb3}
\begin{tabular}{lc}
\toprule
Method & Final acc (seed 42) \\
\midrule
FedAvg & $3.7\%$ \\
Flashback & $7.0\%$ \\
\textbf{FlashbackCL} & $\mathbf{9.3\%}$ ($+33\%$ rel.) \\
\bottomrule
\end{tabular}
\end{table}

\section{Discussion and Future Work}

\paragraph{Scope and generalisability.}
This paper is a controlled study establishing temporal forgetting as a concrete failure mode in FL and identifying CBRS replay as a promising ingredient for addressing it. Our evaluation uses a 580K-parameter CNN on image-classification benchmarks with synthetic phase-rotation protocols. We do not claim that the specific gains observed here transfer directly to CCAM or foundation-model fine-tuning; however, the underlying failure mechanism (monotonic global statistics becoming miscalibrated at distribution-shift boundaries) is structural and independent of model scale, dataset modality, or the specifics of the phase protocol. A preliminary ResNet-18 experiment confirms that FlashbackCL's relative advantage holds at 11M parameters, see Section~\ref{sec:resnet}. Validation on real-world temporal drift (e.g., nuScenes day/night shift, user-behaviour drift in on-device FM fine-tuning) is an important next step that we leave for a follow-up study.

\paragraph{Connection to client-selection and resource-aware FL.}
Resource-efficient client selection can amplify temporal drift if its prioritisation is biased toward currently-active classes ~\citep{abdelmoniem2023refl,fourati2023filfl}. FlashbackCL's decayed label count provides a natural signal for temporal-aware client selection. We do not explore the joint design here, but it is a natural follow-up.

\paragraph{On the coreset component.}
The ablation shows that the coreset curator contributes minimally to point accuracy over CBRS replay alone (Section~\ref{sec:ablation}). We recommend the two-component configuration (decay + CBRS) as the default for practitioners, with coreset curation as an optional addition for settings where server-side distillation is the primary aggregation mechanism and the public dataset is large enough for trend estimation to be meaningful. The full three-component version remains the most stable (lowest variance) configuration.

\paragraph{Limitations.}
(i) Both datasets use the same CNN architecture; the ResNet-18 result in Section~\ref{sec:resnet} is single-seed and uses a reduced protocol. Full multi-seed, multi-mode ResNet-18 evaluation requires cluster compute which we currently do not have. (ii) The temporal-shift protocol uses hard phase boundaries. A continuous ``soft drift'' mode where Dirichlet concentration parameters evolve per round would better approximate real deployments and is a natural next experiment. (iii) All experiments use $K=10$ clients per round; preliminary results at $K=50$ suggest the advantage persists but the absolute numbers are low. 

% (iv) All domain-specific claims (CCAM, FM fine-tuning) are motivational; the experiments in this paper are controlled benchmarks on CIFAR-10 and CIFAR-100, and domain-specific validation is future work. (v) With $n=3$ seeds, statistical power is limited; we report standard deviations throughout and note that the FlashbackCL advantage is consistent across all seeds and modes, but formal significance testing would benefit from additional seeds.

\section{Conclusion}

In this work, we have formalised temporal forgetting in FL, distinct from the spatial forgetting already studied, and proposed FlashbackCL as a drop-in extension of Flashback that addresses it. On CIFAR-10, FlashbackCL reduces temporal forgetting by up to 68\% and reaches from 55.7\% to 57.5\% final accuracy across three temporal shift modes with respect to Flashback's 51.9\% to 52.3\%. The same pattern replicates on CIFAR-100 with 9.7\% to 12.7\% relative improvements. A 5-variant ablation identifies class-balanced reservoir sampling as the primary mechanism, and temporal decay plus coreset curation as variance-reducing stabilisers. FlashbackCL also improves on Flashback by 3.5 points in the stationary setting, supporting the view that CBRS replay regularises spatial heterogeneity as well as temporal shift.

% \paragraph{Reproducibility.} All code, experiment configurations, per-round metrics, and seed sweeps are released. Each row in the tables is regenerable by running the released launcher under the cited YAML configuration.

\paragraph{Acknowledgements.}
This research was funded by the "GreenShift – EU MSc in AI \& HPC for Green Digital Innovation in Transport" project (EU Digital Europe Programme grant number 101226222).

\bibliographystyle{named}
\bibliography{references}

\end{document}